\documentclass[11pt,a4paper]{article}
\usepackage[hyperref]{emnlp-ijcnlp-2019}
\usepackage{times}
\usepackage{latexsym}
\usepackage{graphicx}
\usepackage{amsmath}
\usepackage{amssymb}
\usepackage{enumitem}
\setitemize{noitemsep,topsep=0pt,parsep=0pt,partopsep=0pt}

\usepackage{url}

\aclfinalcopy % Uncomment this line for the final submission

\title{Solving Math Word Problems with Double-Decoder Transformer}

\author{Yuanliang Meng\\
        Text Machine Lab for NLP\\
        Department of Computer Science \\
        University of Massachusetts Lowell \\
        {\tt ymeng@cs.uml.edu} \\\And 
        Anna Rumshisky \\
        Text Machine Lab for NLP\\
        Department of Computer Science \\
        University of Massachusetts Lowell \\
        {\tt arum@cs.uml.edu} }

\date{}

\begin{document}
\maketitle
\begin{abstract}
    This paper proposes a Transformer-based model to generate equations for math word problems. It achieves much better results than RNN models when copy and align mechanisms are not used, and can outperform complex copy and align RNN models. We also show that training a Transformer jointly in a generation task with two decoders, left-to-right and right-to-left, is beneficial. Such a Transformer performs better than the one with just one decoder not only because of the ensemble effect, but also because it improves the encoder training procedure. We also experiment with adding reinforcement learning to our model, showing improved  performance compared to MLE training.
\end{abstract}

\section{Introduction}
Automatically constructing formulae and equations to solve math word problems is a challenging task for artificial intelligence. The attempts to solve it started as early as the 1960s~\cite{Bobrow:1964:NLI:889266}, and in the 1980s some efforts were made to model the cognitive process of humans~\cite{briars_larkin_1984,fletcher_1985}. More recently, statistical machine learning methods have been adopted~\cite{P14-1026,D15-1096,Q15-1042,P16-1084}. These models use template matching and only work for highly restricted types of problems. Template matching compares input problems with the problems in training set, and then uses the equation template from the best match.

With the development of deep learning techniques in NLP, RNN-based sequence-to-sequence models have been used to generate equations~\cite{D17-1088,P17-1015}. So far their performance is not as good as template matching, although they are able to generate correct equations that do not have exact copies in training set. 
\newcite{C18-1018} makes a hybrid model, which conducts template retrieval when good candidates are found, and generates new equations with GRU if there is no good candidate. Their results are still the state-of-the-art on a number of datasets, including their own Dolphin18K, which is a relatively big dataset covering a large variety of math word problems. 

RNN-based models have two major issues. First of all, they tend to over-generate number tokens. 
%For example, if a math problem has two numbers $N_1$ and $N_2$, RNN often generates extra tokens e.g. $N_3$ in the equations. 
Secondly, they often put numbers in wrong positions in an equation, even if the equation has the correct shape. 
%\newcite{C18-1018} implemented copy and align modules in the neural network system to alleviate the problems.
%Before 2017, the most popular sequence transduction NLP models are based on RNNs, such as LSTM and GRU, often incorporated with attention mechanisms. 
The recently introduced Transformer~\cite{Transformer} model has shown improved performance in a number of NLP tasks. It removes the recurrence or convolution of deep neural networks, but relies on richer attention mechanisms. %Specifically, each position in the encoder can attend to all positions in the previous layer; and each position in the decoder attends over all positions in the input sequence.
In the original Transformer model, the decoder is unidirectional, typically generating tokens left to right, conditioning on previously generated output.  
However, it is quite possible that in some cases attending to previously generated output in the opposite direction may in fact generate better output.
%
%However the decoder may attend to encoder better when the direction is not left to right.

Based on this intuition, in this work, we generate equations using Transformer with two decoders working in opposite directions. \footnote{Note that BERT model~\cite{DBLP:journals/corr/abs-1810-04805} which came out when the present work was under way, follows a similar intuition, conditioning on both left and right context; however, it can not be adopted for generation directly, since the output typically needs to be generated sequentially.}
%In this paper, we generate equations using Transformer with two decoders in both directions (left-to-right and right-to-left).
%
%
%Bidirectional Encoder Representations from Transformers (BERT)~\cite{DBLP:journals/corr/abs-1810-04805} model uses bidirectional Transformer encoders to construct a language model representations with left and right context. However, sequence generation tasks require generating tokens one by one, and new tokens are only conditioned on previously generated tokens. 
%Therefore the architecture of BERT cannot be adopted directly into sequence-to-sequence models. 
%Therefore the decoders have to follow some directionality, unlike building a language model.
% AR: BERT, which came out after this work was started, had followed a similar intiutition that unidirectional decoder is not sufficient; however, BERT model can not be used for generation...
%
In this paper, our major contributions are the following: 
\begin{enumerate}
\setlength{\itemsep}{0pt}
    \setlength{\parskip}{0pt}
    \setlength{\parsep}{0pt} 
    \item This work is the first to use Transformer to generate mathematical equations. Note that compared to natural language, equations have a very low tolerance to grammatical errors, but allow multiple equivalent forms.
        %We use Transformer to generate mathematical equations, which are quite different from natural language text. As far as we know, it is the first attempt to do so.
    \item We show two Transformer decoders working in opposite directions work better than a single decoder. When the encoder is trained with loss from both decoders, the performance is better overall, regardless of whether the output is generated by both decoders in an ensemble, or just a single decoder.
    %This is true \emph{with and without ensemble}, suggesting improved quality of the encoder.
    \item  Our model does not use special copy and align modules to copy over numbers and align them to correct positions, yet it outperforms RNN architectures that use such mechanisms.
    
    %Our model outperforms RNN architectures that use copy and align mechanisms, doing so without using any mechanisms to copy over numbers and align them to correct positions, 
    %Our model outperforms RNN architectures that used 
    %\item  Our model does not have copy and align mechanisms, which are used by some RNN architectures to copy over numbers and align them to correct positions, but can still achieve better performance than RNN-based model with them.
    
    %\item Reinforcement learning can further improve the results.
\end{enumerate}
%
%The code repository can be found on Github~\footnote{\url{https://github.com/text-machine-lab/mathword}}
Our proposed architecture outperforms all models that use the MLE objective, although we are not able to outperform more complicated systems which, e.g., use template retrieval and rely on handcrafted features. With reinforcement learning, the model has better performance than using MLE.
%It should be mentioned that we did not beat the system with template retrieval method, which is not a generation model and relies on handcrafted features.
%
Our code will be made available on GitHub.

\section{Dataset}
We used the Dolphin18k dataset\footnote{\url{https://www.microsoft.com/en-us/research/project/sigmadolphin-2/}}~\cite{P16-1084}. It contains 18,460 problems posted by users on Yahoo! Answers. The dataset has two partitions, the Dev set contains 3,728 problems and the Eval set 14,732 problems. Following the authors' method, final evaluation is performed on the Eval set with 5-fold cross evaluation. Therefore there is no division between training set and test set. %Figure~\ref{fig:example} demonstrates an example of the data.
Here is an example problem:
\begin{figure}[ht]
\begin{center}
%\fbox{
\includegraphics[width=1.0\columnwidth]{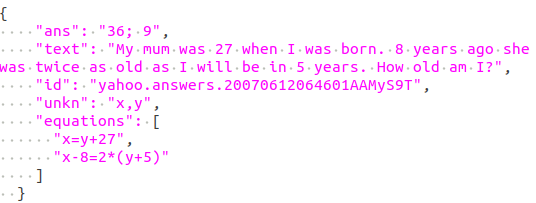}
%}
\end{center}
\vspace{-1.5em}
%\caption{\label{fig:example} Example of a math word problem.}
\label{fig:overview}
\end{figure}

In this paper, we focus on the T6 subset of Dolphin18k. The problems in this subset all have at least 5 similar problems, with similar equation types. There are supposed to be 6,872 problems in this subset, with 1,578 manually annotated and 4,826 automatically annotated. However Yahoo! deleted some of them and 6,762 instances are still available. 
Note that a small portion of them miss either the equations or the answers, or are otherwise erroneous or impossible to solve.
%Over 95\% problems do have equations matching the answer. 
%Some annotated equations are in poor quality, and are useless as training data. For example, they skip some steps and the derivations are not shown. 
%In order to make a fair comparison with other systems, we keep all problems in training and evaluation even if they are erroneous.  
We keep such problems in the data to enable fair comparison with previous work.

%Some problems are solvable but it is impossible to learn to generate equations from the given text. Many of them are problems of permutations and combinations, some others involve series, prime factorization, fraction simplification and so on. These are not our target at this time, although they are also included for a fair evaluation. 

\section{System}
\begin{figure}[h]
    %\centering
    \includegraphics[width=1.0\columnwidth]{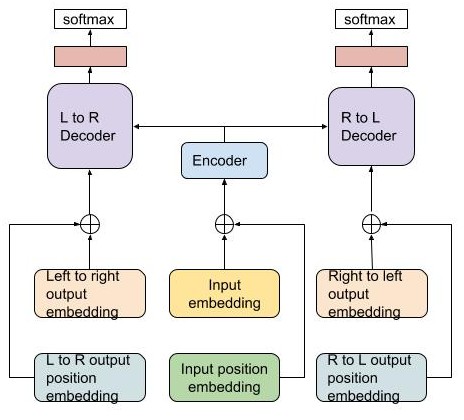}
    \caption{Modified Transformer with two decoders: left-to-right and right-to-left.}
    \label{fig:system}
\end{figure}

The backbone of the system is a sequence-to-sequence model based on Transformer with two decoders which generate output sequences in both directions (left-to-right and right-to-left). As a standard Transformer, it uses token embeddings and position embeddings as input. Each decoder has its own input (equation token embeddings and position embeddings), read in the corresponding direction 
(see Figure~\ref{fig:system}).

Multiple equations are separated with ``;''.
%The output can be one or multiple equations. If multiple, equations are separated with ``;'', but they are still in the same sequence. %Token embeddings for input and output are separate. 
Number tokens are converted to special symbols, both in text and in equations. The Transformer is trained to generate equations with those symbols without seeing actual values. Symbols are converted back to their original values when the equations are solved. The solutions are then compared to key answers for evaluation.

\subsection{Mapping number tokens}
%Numbers are infinite so they must be 
We represent numbers with fixed-size tokens. \newcite{D17-1088} and \newcite{C18-1018} use a list of number tokens $\{N_1, N_2,...,N_m\}$ to represent numbers. We slightly modify the idea and introduce three types of tokens: negative numbers, float between (0, 1), and others. 
%For the given set of problems, negative numbers are not often used in the text. Decimals between (0, 1) often represents ratios, percentages and so on, so it is better to separate the symbols.
\iffalse
In the example
\begin{quote}
    "Find the equation of the line that goes through the points (-15, 70) and (5, 10)?"
\end{quote}
We will change the tokens to ``Find the equation of the line that goes through the points $(M_1, N_2)$ and $(N_3, N_4)$?" Here $M_1$ shows it is the first number token and it is negative. Note the index of the number tokens reflects their positions in the text, and the type has no effect. In the math word problems, negative numbers are not used often but it often needs special treatment in equation formatting. For example, it must be put in parenthesis when it follows an operator. Decimals between (0, 1) often represents ratios, percentages and so on, so it is better to separate the symbols.  
\fi

In order to obtain valid training data, it is crucial to make sure the number tokens in text (input) can properly align with number tokens in equations (output). In practice, this is difficult because number format may be inconsistent.
%the numbers in the files often have inconsistent formats. 
For example, 
%the problem text may have $25\%$, but the corresponding equation may use $0.25$ or even just $/4$ somewhere. Fractions are particularly troublesome. 
 $3~1/3$ in text may correspond to $3.33$ or $10/3$ in the equation. We try all  possibilities to find a match. 
 %If the input text has $3~1/3$ and the equation has something like $3+1/3$, then we identify two tokens. On the other hand, if the equation has $3.333$, we convert the input $3~1/3$ to one token. 

%Another problem is different number tokens may happen to have the same value.
%Different number tokens may also have the same values. 

Mapping number tokens from equation to text is nontrivial in general, since different number tokens may have the same value.
%
%the text may have identical numbers, but they should be assigned with different number tokens $N_i$ and $N_j$ to make equations general. 
%An arbitrary alignment between symbols and values in test data may not cause a problem, but in training data it leads to undesirable equations which cannot generalize. Fortunately, the problem text tends to avoid identical values. 
Generally speaking, we assign the symbols in order i.e. if $N_1$ and $N_2$ both have the same value $3$, then in the equations, we convert the first occurrence (left to right) of $3$ to $N_1$ and the second to $N_2$. %To tackle the problem in a better way, there needs to be an independent alignment model, which we consider unnecessary. 

%As mentioned before, the annotated equations are often not in good format, which makes learning impossible in some cases. %For instance, sometimes two numbers are added but only the sum shows up in equation, without any derivation demonstrated. For now we want to keep the data that way in order to make a fair comparison with other models. 
%Presumably, rewriting the equations can significantly improve results.

\subsection{Transformer model}
%The Transformer has an encoder allowing mutual attention between representations of all pairs of positions (time steps). Its decoder also allows attention between input and output in each position. Such a model has been proved to be successful in a number of tasks such as machine translation and constituency parsing. However, they are all in natural language domain, and our work needs to address mathematical domain.

Mathematical equations and natural language share some properties, since both use sequences of symbols to represent meaning. However, equations have a more restricted vocabulary with synonyms explicitly prohibited. Equations also have a low tolerance for grammatical errors. While a slightly ill-formed sentence is comprehensible, an ill-formed equation has no solution and is useless. On the other hand, each equation has a number of mathematically equivalent forms, and our training data does have such variance. %Natural language can also have multiple expressions with the same meaning, but usually there are subtle semantic or pragmatic differences.%, so they are seldom truly equivalent given a context.

Because of the differences, it is instructive to experiment with the Transformer model in the equation domain.
%we must conduct experiments to see if Transformer truly works in mathematical domain.
Our encoder resembles the canonical Transformer encoder, but we use two decoders. One is decodes left to right and the other right to left. These two decoders are jointly trained with the shared encoder. 
Some of the motivations for this choice are as follows.  First, as mentioned above, having two decoders going in opposite directions improves the training of the encoder; the intuition here is similar to the masked language model in BERT, which benefits from having both left and right context.  Second, the L-to-R and R-to-L decoder can each cast a vote according to their confidence score, creating an ensemble decision.

%since Transformer  uses mutual attention between input and output, with two decoders, the encoder can be more properly trained, similar to training the masked LM in BERT, where the language model is trained with all possible contexts and is strictly more powerful than a left-to-right model.
%Moreover, (2) outputs of the two decoders can cast a vote according to their confident scores and make an ensemble. 

%There are at least two motivations for using two decoders. (1) Since Transformer has mutual attention between input and output, with two decoders the encoder can be more properly trained, similar to training the masked LM in BERT. Moreover, (2) outputs of the two decoders can cast a vote according to their confident scores and make an ensemble. 

As most sequence-to-sequence models, we use cross-entropy (CE) loss as the objective, which is equivalent to maximum likelihood estimation (MLE). As shown in Equation~\ref{equ:loss}, the CE losses from the two decoders are added for training.
\begin{align}\label{equ:loss}
    L_{CE}=&-\sum_{t=1}^{T}\textup{logP}_\theta(y_t|y_{0:t-1}, X) \nonumber \\ 
     &-\sum_{t=0}^{T-1}\textup{logP}_\theta(y_t|y_{t+1:T}, X)
\end{align}
For prediction, each decoder generates its own equations. We use the log probability scores to perform a beam search. After obtaining the top results from the left-to-right and right-to-left beams, respectively, we pick the one with the higher score as the final result.
\iffalse
\begin{equation}\label{inf}
    y_{pred}=\underset{\widehat{y}^i}{argmax}\sum_{t=1}^{T}\textup{logP}_\theta(\widehat{y}^{i}_t|\widehat{y}^{i}_{1:t-1}, X) 
\end{equation}
Once the model is trained, each decoder generates a beam of sequences. %following equation~\ref{inf}. 
%Here $i$ is the index of a candidate sequence, and $\widehat{y}^{i}_t$ represents the prediction at time $t$ from the $i$-th candidate sequence. 
%The sequence with the highest score from the beam will be selected. Then the two sequences from the decoders are compared, and we use the one with a higher score.
\fi
%
We encode token positions with sinusoids in the same way as~\newcite{Transformer}. Output positions also have two directions, each for the corresponding decoder. 

\subsection{Reinforcement learning}
Sequence-to-sequence models with minimum CE loss has the \emph{exposure bias}~\cite{DBLP:journals/corr/RanzatoCAZ15} problem. For training, ground truth tokens are fed into decoder, but predicted tokens are used during testing.
%A compromise is is to use \emph{schedule sampling}~\cite{Bengio:2015:SSS:2969239.2969370} i.e. use teacher forcing for some epochs and gradually replace ground truth with model generation. 
In addition to that, there is a disparity between evaluation metrics and training objective. The ultimate goal is to maximize the number of correct answers via solving generated equations, not to generate exact copies of equations in training data. For this purpose, a natural way is to use reinforcement learning. We can set a positive reward $r=1$ when a correct solution is found from the equation, and $r=0$ if the solution is incorrect or absent. 
\iffalse
\begin{equation}
    r=\left\{\begin{matrix}
1, \text{~if answer is correct}\\ 
0, \text{~otherwise}
\end{matrix}\right.
\end{equation}
\fi

The objective is to maximize the expected reward, given the distribution of output sequences. $\pi_{\theta}$ represents the trained policy function. 
%In this case, ideally, it generates the optimal output sequences based on input sequences and its internal states. 
In order to optimize the policy, we use the \emph{REINFORCE} algorithm~\cite{Williams:1992:SSG:139611.139614}. It can be proved that optimizing the policy with respect to the expected reward is equivalent to equation~\ref{sample_loss}.
\begin{equation}
\begin{split}\label{sample_loss}
    L_{\theta} &=-\mathbb{E}_{\widehat{y}_{1:T}\sim \pi_{\theta}(\widehat{y}_{1:T})}\textup{log}\pi_{\theta}(\widehat{y}_{1:T})\times r(\widehat{y}_{1:T}) \\
&\approx -\frac{1}{N}\sum_{n=1}^{N}\sum_{t=1}^{T}\textup{log}\pi_{\theta}(\widehat{y}_t|\widehat{y}_{t-1})\times [r(\widehat{y}_{1:T})-r_b]
\end{split}
\end{equation}
In practice, it is very expensive to fully simulate the distribution of output and we can just obtain a few output sequences using beam search, and draw $N$ samples from them. The baseline $r_b$ controls relative gradient amplitudes between high-reward results and low-reward results. We choose it as the mean reward among $N$ samples.
\begin{equation}
    r_b=\frac{1}{N}\sum_{n=1}^{N}r_n(\widehat{y}_{1:T})
\end{equation}

The \emph{REINFORCE} algorithm tends to be very slow and often hard to converge. We always train models with cross-entropy loss first, and then continue to train with \emph{REINFORCE}.

\section{Experiments}
For both the minimum cross-entropy learning and the reinforcement learning, hyperparameters are tuned on the validation set of Dolphin18KT6, and final evaluation is performed on the evaluation set with non-repeated 5-fold cross validation. 
We use SymPy\footnote{https://www.sympy.org/en/index.html} to solve the equations and obtain accuracy scores. If an equation is ill-formed or cannot be solved, it is considered a wrong solution.

\subsection{Hyperparameters and configurations}
Word embeddings for question text are initialized with \textsf{glove.840B.300d} word vectors\footnote{https://nlp.stanford.edu/projects/glove/}. Each of them has 300 dimensions. The encoder and decoder each has 3 layers. Because the linear layers in Transformer has the dimensionality of 512, an linear layer with 512 output units is added right after the embedding layer. Other hyperparameters are the same as in \newcite{Transformer}. The beam size for reinforcement learning sampling is 6. For prediction, the beam size is 10.

The minimum cross-entropy model is trained with 120 epochs. We use the Adam optimizer with the learning rate of 5e-5, $\beta1$ = 0.9, $\beta2$ = 0.98.
Reinforcement learning training is implemented after that with the learning rate 5e-7. %More details of the settings can be found in our repository. 

\subsection{Results}

\begin{table}[t]
\begin{center}
\resizebox{\columnwidth}{!}{
\begin{tabular}{|l|c|}
\hline 
\bf{Generative Models - RNN} &\bf{Accuracy} \\
\hline
GRU w/ Attn (MLE) &13.0   \\
+Copy+Align (MLE) &21.0 \\
+Copy+Align (RL) &23.3 \\
\hline
\hline
\bf{Generative Models - Transformer} &\bf{Accuracy} \\
\hline
1-Decoder Transformer (MLE) & 19.4\\
2-Decoder Transformer, L-to-R (MLE) & 20.6 \\
2-Decoder Transformer, R-to-L (MLE) & 20.2 \\
2-Decoder Transformer, vote (MLE) &21.7 \\
2-Decoder Transformer, vote (RL) &22.1 \\
\hline
\hline
\bf{Retrieval Models} &\bf{Accuracy} \\
\hline
\newcite{D17-1084} &30.6  \\
(Hybrid) +GRU+Copy+Align (RL) &33.2  \\
\hline
\end{tabular}
}
\end{center}
%\vspace{-1.5em}
\caption{\label{tab:results} Results of RNN models and retrieval models are from \newcite{C18-1018}. Results of Transformer models are from our experiments. Our Transformer models do not use copy or align. MLE stands for maximum likelihood learning. RL stands for reinforcement learning.}
\end{table}

Table ~\ref{tab:results} shows all the results from our system as well as the state-of-the-art. Results of the RNN models and retrieval models are from \newcite{C18-1018}. When using the MLE objective, we can see that Transformer outperforms GRU, especially when it has no copy and align mechanisms. Another important observation is that the results with two decoders outperform the one with one decoder by a large margin. Even if we just pick the results from one decoder and ignore the other, the two-decoder model still yields better results. It suggests that the whole system (including the encoder) benefits from having two decoders.

We have not been able to outperform template retrieval models. These models use input problems as queries, and search in the training dataset to find the best match. Then the equations of the best match serve as the template, and another module fills in the  number values in the template. Obviously, such method can only solve problems which have very similar examples in training set, but it can be robust, depending on the dataset. They hybrid system tries template retrieval first, and uses the RNN model if there is no good candidate. With reinforcement learning, our model shows improved  performance, but the improvement does not seem to be as big as the RNN+Copy+Align achieved with RL. Additional experiments with dual decoder Transformer model with RL may be needed to establish whether copy-and-align mechanisms are necessary.
%The reason is unclear so far. Usually RL is hard to converge and has big variances, and probably we did not find the best model configuration. 

As mentioned before, the equations in the dataset are often not in good format. Some times they contain values that are not directly found in the problem text, and do not show the derivation. If the equations can be rewritten in a better way,  sequence-to-sequence models will likely show improved performance.
%be trained more properly.

\section{Conclusion}
We have shown that a Transformer-based model can generate equations for math word problems, and it has an edge over RNN-based models. Jointly training two decoders with a shared encoder in the Transformer works better than using just one decoder. This is true even without the ensemble effect. Reinforcement learning can further boost performance.
So far the template retrieval method still beats generative models. This is partly due to the low quality of ground truth equations. If the derivation of equations are annotated in a better way, the generative models may be able to learn better.

\iffalse
\section*{Acknowledgments}

The acknowledgments should go immediately before the references.  Do
not number the acknowledgments section. Do not include this section
when submitting your paper for review. \\
\fi

\bibliography{emnlp-ijcnlp-2019}
\bibliographystyle{acl_natbib}

\iffalse
\appendix

\section{Appendices}
\label{sec:appendix}

An example of a math word problem. The text can be mapped to two linear equations. In our model the numbers $27, 8, 2, 5$ are converted to $N_1, N_2, N_3, N_4$ so the equations can be generalized as a training instance.
\begin{figure*}[h]
\begin{center}
%\fbox{
\includegraphics[width=1.0\columnwidth]{math_example.png}
%}
\end{center}
\vspace{-1.5em}
%\caption{\label{fig:example} Example of a math word problem.}
\label{fig:overview}
\end{figure*}

\fi

\end{document}